\def\year{2020}\relax
\documentclass[letterpaper]{article} 
\usepackage{aaai20}  
\usepackage{times}  
\usepackage{helvet} 
\usepackage{courier}  
\usepackage[hyphens]{url}  
\usepackage{graphicx} 
\usepackage{graphicx}  
\usepackage{amsfonts}
\usepackage{makecell}
\usepackage{enumerate}
\title{Goal-Oriented Multi-Task BERT-Based Dialogue State Tracker}
\author{Pavel Gulyaev, Eugenia Elistratova, Vasily Konovalov, \AND Yuri Kuratov, Leonid Pugachev, Mikhail Burtsev  \\ \\
Neural Networks and Deep Learning Lab \\
Moscow Institute of Physics and Technology \\
Institutskiy Pereulok, 9 \\
Dolgoprudny, Moscow Oblast, 141701 \\
guliaev.pa@phystech.edu 
}

\begin{document}
\maketitle
\begin{abstract}
Dialogue State Tracking (DST) is a core component of virtual assistants such as Alexa or Siri. To accomplish various tasks, these assistants need to support an increasing number of services and APIs. The Schema-Guided State Tracking track of the 8th Dialogue System Technology Challenge highlighted the DST problem for unseen services. The organizers introduced the Schema-Guided Dialogue (SGD) dataset with multi-domain conversations and released a zero-shot dialogue state tracking model. In this work, we propose a \textbf{GO}a\textbf{L}-\textbf{O}riented \textbf{M}ulti-task \textbf{B}ERT-based dialogue state tracker (GOLOMB) inspired by architectures for reading comprehension question answering systems. The model “queries” dialogue history with descriptions of slots and services as well as possible values of slots. This allows to transfer slot values in multi-domain dialogues and have a capability to scale to unseen slot types. Our model achieves a joint goal accuracy of 53.97\% on the SGD dataset, outperforming the baseline model.
\end{abstract}


\section{Introduction}

The advent of virtual assistants such as Amazon Alexa, Google Assistant and many others resulted in an urge to develop applications providing a natural language interface to services and APIs. These task-oriented dialogue systems can be implemented by either knowledge-based or data-driven approaches. Dialogue state tracking (DST) is the main component in task-oriented dialogue systems. DST is responsible for extracting the goals of the user and the (slot, value) pairs corresponding to them. For example, if the user's goal is to order a taxi (intent: \textit{OrderTaxi}), then the slots are \textit{destination}, \textit{number\_of\_passengers} etc. The development of task-oriented dialogue systems has been put forward by releases of task-oriented dialogue corpora such as NegoChat \cite{konovalov2016negochat}, ATIS \cite{hemphill1990atis} and many others. However, these single-domain datasets do not fully represent the challenges of the real world, where a conversation often revolves around multiple domains.

The release of the multi-domain dialogue dataset (Multi-WOZ) raised the bar in DST due to its mixed-domain conversations \cite{eric2019multiwoz}. This dataset contains dialogues where the domain switches over time. For example, a user might start a conversation by asking to reserve a restaurant, then go on to request a taxi ride to that restaurant. In this case, the DST has to determine the corresponding domain, slots and values at each turn of the dialogue, taking into account the history of the conversation if necessary.

The largest public task-oriented dialogue corpus, the Schema-Guided Dialogue (SGD) dataset, has been recently released by Google \cite{rastogi2019towards}. It contains over 16,000 dialogues in the training set spanning 26 services in 16 domains. In addition, to measure the model’s ability to perform in zero-shot settings, the evaluation sets (dev and test) contain unseen services and domains.

SGD provides a schema for each service represented in the dialogues. A schema is a list of slots and intents for the service accompanied by their natural language description. The dialogue state consists of three fields: \texttt{active\_intent}, \texttt{requested\_slots} and \texttt{slot\_values}. SGD encourages the model to recognize the semantics of intents and slots from their descriptions while predicting the dialogue state, enabling zero-shot generalization to new schemas. The authors also proposed a single unified task-oriented dialogue model for all services and APIs, achieving 41.1\% joint goal accuracy when trained and evaluated on the entire dataset and 48.6\% joint goal accuracy for single-domain dialogues only. The proposed model encodes each schema element (intents, slots and categorical slot values) using its natural language description provided in the schema file. These embedded representations are not fine-tuned afterwards, which seems to be a disadvantage of the suggested approach.

In this paper, we introduce a \textbf{GO}a\textbf{L}-\textbf{O}riented \textbf{M}ulti-task \textbf{B}ERT-based dialogue state tracker (\textbf{GOLOMB}) inspired by recent reading comprehension question answering neural architectures such as \cite{devlin2018bert}. These models search the text for a span that contains the answer to the user’s question. We reformulate the dialogue state tracking task to have a similar form. Given a dialogue history and a “question”, comprising the slot, service and intent descriptions, the model should return values for a dialogue state as an “answer”. To predict a dialogue state update, our model solves several classification tasks and a span-prediction task. For each of these tasks, there is a special head implemented as a fully connected linear layer. This architecture makes it possible to jointly train the representations for schema elements and dialogue history. Our approach is robust to changes in schema due to zero-shot adaptation to new intents and slots. In addition, the proposed model does not rely on a pre-calculated schema representation. GOLOMB outperforms the baseline and achieves a joint goal accuracy of 53.97\% on the dev set. The model is publicly available \footnote{\url{https://gitlab.com/zagerpaul/squad_dst}}.


\section{Related Work}

The main task of dialogue state tracking is the identification of all existing slots, their values and the intentions that form them. The pairs of slots and their values form the state of the dialogue. The dialogue state defines the interaction with the external backend API and the selection of the system’s next action.

Classic dialogue state tracking models combine the semantics extracted by natural language understanding module with the previous dialogue context to estimate current dialogue state \cite{thomson2010bayesian,wang2013simple,williams2014web} or jointly learn speech understanding and dialogue tracking \cite{henderson2014word,zilka2015incremental,wen2016network}. In past tasks, such as DSTC2 \cite{henderson-etal-2014-second} or WoZ \cite{wen2016network}, it was required to track the dialogue within one domain. At the same time, all possible values for all slots were given in the dataset ontologies. Thus, the dialogue state tracking task was reduced to enumerating and then selecting pairs of slot values. This results in ontology specific solutions unable to adapt to new data domains. For example, the Neural Belief Tracker uses word representation learning to obtain independent representations for each slot-value pair \cite{mrkvsic2017neural}. Developers of Global-Locally Self-Attentive Dialogue State Tracker (GLAD) found that 38.6\% of turns in the WoZ dataset contain rare slot-value pairs with fewer than 20 training examples \cite{zhong2018global}. There is therefore not enough training data for many of the slots, which greatly decreases joint goal accuracy. To solve this problem, the authors of GLAD proposed sharing parameters between all slots. Thus, information extracted from some slots can be used for other slots during training, which increases the quality of state tracking and makes it possible to work with multi-domain dialogues. However, the model uses both the parameters common to all slots and the parameters trained individually for each slot.

As technology for dialogue state tracking developed, a more complex task was proposed in the MultiWoZ dataset \cite{budzianowski2018multiwoz,eric2019multiwoz}. Here, the system needs to extract a state from dialogues where the user can switch between domains or even mention multiple domains at the same time. As the number of possible slots and their possible values grew, iterating over all pairs became labor-intensive and learning slot-specific parameters became less efficient.

The Globally-Conditioned Encoder (GCE) \cite{nouri2018toward} is an improved version of GLAD. This model, in which all parameters were shared between all slots, surpassed the previous model for WoZ and MultiWoZ tasks. StateNet \cite{ren2018towards} generates a representation of dialogue history and compares it to the slot value representation in the candidate set. Here, the dialogue history consists of a system’s act and the subsequent user utterance. The HyST model \cite{goel2019hyst} forms a representation of the user’s utterances with hierarchical LSTM, and then combines two approaches for selecting slot values. The first one independently estimates the probability of filling the slot by each candidate from the candidate set. The second estimates the probability distribution over all the possible values for that slot type.

The majority of the aforementioned models require a vocabulary with all the values supported by the model. Thus, it is not possible to process out-of-vocabulary values. To address this issue, the PtrNet \cite{xu2018end} model uses an index-based pointer network for different slots. The TRADE model \cite{wu2019transferable} tracks the dialogue state using a biGRU-based encoder and decoder. The encoder encodes each token in the dialogue history. The decoder generates slot value tokens using a soft-copy mechanism that combines attention over the dialogue history and value selection from the vocabulary. The authors also studied zero- and few-shot learning to track the state of the out-of-domain dialogues. Also, pre-trained language models can help with handling unknown values and zero-shot learning. BERT-DST \cite{chao2019bert} uses BERT to predict the start and the end tokens of the value span for each slot.


\section{GOLOMB Model}

\begin{figure*}[t!]
\centering
\includegraphics[width=2\columnwidth]{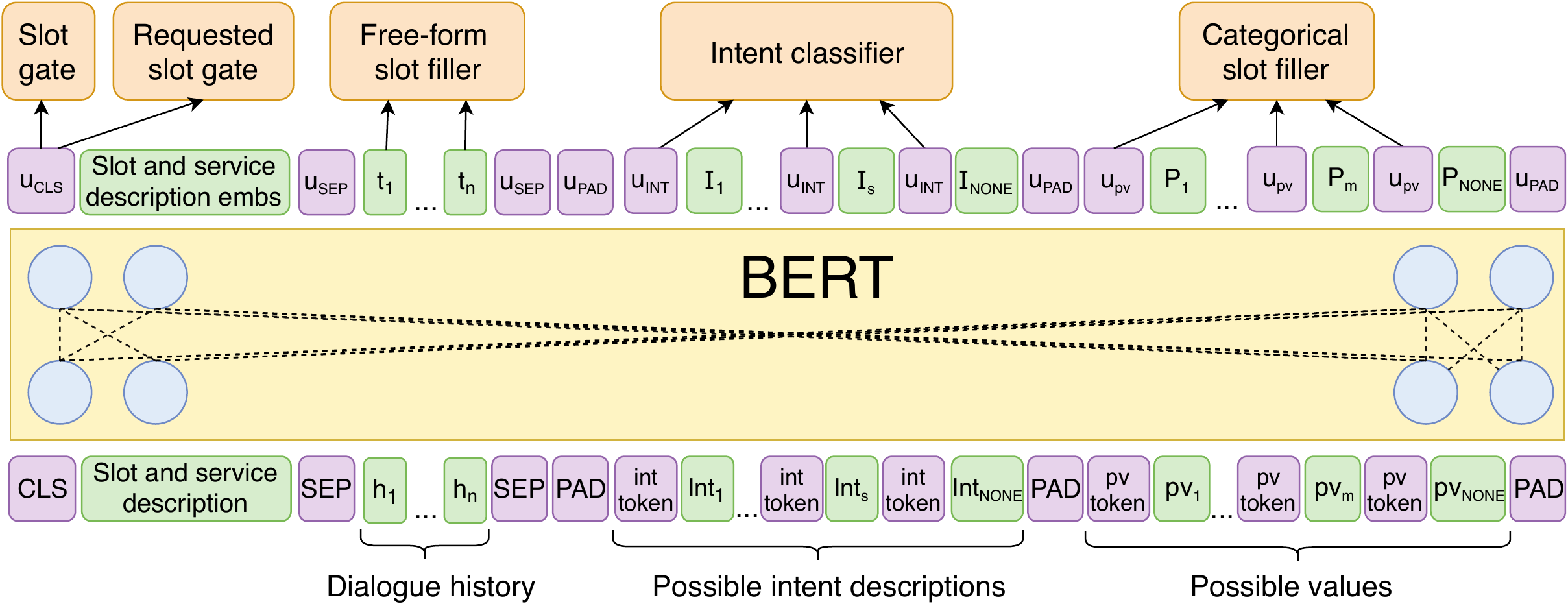}
\caption{The architecture of \textbf{GO}a\textbf{L}-\textbf{O}riented \textbf{M}ulti-task \textbf{B}ERT-based dialogue state tracker (\textbf{GOLOMB}). The \textit{slot gate} head is used to decide whether a slot has to be included in the final state. The \textit{requested slot gate} predicts whether a slot has been requested by the user. The \textit{intent classifier} head chooses the active intent. Depending on whether the slot is categorical or non-categorical, different heads are used. For a non-categorical slot, the \textit{free-form slot filler} selects positions of the beginning and the end of the slot value in the dialogue history. For a categorical slot, the \textit{categorical slot filler} selects the slot value among the possible values.
}
\label{PaperModel}
\end{figure*}

In this section, we provide a detailed description of the proposed GOLOMB model (see Figure \ref{PaperModel}). The input of the model consists of a slot description and a dialogue history followed by the possible slot values and the supported intent descriptions. A BERT-based encoder converts the input into contextualized sentence-level and token-level representations. These representations are then fed into the following task-specific output heads:

\begin{enumerate}
    \item Classification heads:
    \begin{itemize}
        \item \textbf{Intent classifier} is responsible for active intent prediction.
        \item \textbf{Requested slot gate} predicts the list of slots requested by the user in the current turn.
        \item \textbf{Slot gate} predicts whether a slot is presented in the context.
        \item \textbf{Categorical slot filler} performs slot value prediction by selecting the most probable value from the list specified in the schema.
    \end{itemize}
    \item Span-prediction head:
    \begin{itemize}
        \item \textbf{Free-form slot filler} performs a slot value prediction by identifying it as a span in the context.
    \end{itemize}
\end{enumerate}

Each head is implemented as a fully connected linear layer.

\subsection{Schema-Guided Dialogue Task}

The dialogue state in the Schema-Guided Dialogue dataset is a frame-based representation of the user’s goals retrieved from the dialogue context. It is used to identify an appropriate service call and assign the values of the slots required for that call. The dialogue state consists of \textit{active\_intent}, \textit{requested\_slots} and \textit{slot\_values}.

A dialogue in the SGD dataset is represented by a sequence of turns between a user and a system. The turn annotation is organized into frames where each frame corresponds to a single service. A separate dialogue state is maintained for each service in the corresponding frame.
A \emph{state update} is defined as the difference between the slot values present in the current service frame and the frame for the same service for the previous user utterance. The task is state update prediction.

\subsection{BERT Encoder} 

We assemble a group of input sequences for every frame according to its service schema. For each slot in the schema an input sequence is formed and then fed into the pre-trained BERT encoder. We adopt the input structure from the BERT-based model for question answering \cite{devlin2018bert} on the SQuAD dataset \cite{rajpurkar2016squad}, which consists of a question part and a context part. In our case, the context is a dialogue history and the question is a concatenation of slot and domain descriptions. Table \ref{tab:bert_input} shows the components of the input sequence.

    \begin{table}[t]
        \centering
        \resizebox{0.47\textwidth}{!}{%
        \begin{tabular}{c|c}
        \Xhline{1pt}
                                         &  \textbf{Input sequence} \\
             \hline
             \textbf{Question}           & Slot and service description \\
             \textbf{Context}            & Dialogue history \\
             \textbf{Possible intents}   & Descriptions of intents supported by the service \\
             \textbf{Possible values}    & Possible slot values (for categorical slots only) \\
        \Xhline{1pt}
        \end{tabular}}
        \caption{The components of GOLOMB input.}
        \label{tab:bert_input}
    \end{table}
    
The full input sequence is shown at the bottom of the diagram in Figure \ref{PaperModel}. It starts with a $\texttt{[CLS]}$ token followed by the concatenation of the slot and domain descriptions. The next part, separated by $\texttt{[SEP]}$ tokens, is the dialogue history. In our case, it is the current user utterance with the preceding system utterance. We then pad the input until the \texttt{max\_hist\_len} is reached (by default \texttt{max\_hist\_len=250}). After that, we add all relevant intent descriptions separated by the special $\texttt{[int]}$ token and padded to the \texttt{max\_intent\_len} (\texttt{max\_intent\_len=50} by default). Finally, if the slot is categorical, we complete the input with its possible values $pv_i$ accompanied by the special token \texttt{[pv]}. We also add the special values “NONE” to the possible intent and slot values so as not to penalize the model if the intent or slot is not present in the context. For every input token, the BERT-based encoder generates a contextualized embedding. Different parts of the encoder output are read out by different heads (see Figure \ref{PaperModel}).

\subsection{GOLOMB Heads}

All heads perform a linear transformation of the corresponding embeddings. Let $x$ be a vector from $\mathbb{R}^n$ and let $m$ be an arbitrary positive integer. Then, for head $T$, $\mathcal{F}_{T, m}: \mathbb{R}^n \to \mathbb{R}^m$ is a projection that transforms $x$ into the prediction vector $y \in \mathbb{R}^m$:
\begin{equation}
    \mathcal{F}_{T, m}(x) = y,
\end{equation}
where $\mathcal{F}_{T, m}$ is implemented as a single fully connected layer without an activation function:
\begin{equation}
    \mathcal{F}_{T, m}(x) = W_{T, m} x + b_{T, m}.
\end{equation}

\subsubsection{Slot gate}

For each slot, the model predicts its values, but not all slots should be included in the state update. The Slot gate head predicts the slot status, which can have three values: \texttt{ptr}, \texttt{dontcare} and \texttt{none}. If the slot status is predicted to be \texttt{none}, then this slot and its value will not be included in the state update. If the prediction is \texttt{dontcare}, then the special value \texttt{dontcare} is assigned to the slot. If the slot status is \texttt{ptr}, the slot value predicted by one of the slot fillers will be included in the state update.

The slot status is obtained by applying $\mathcal{F}_{\mathrm{status}, 3}$ to the $\mathrm{u_{CLS}}$ embedding:
\begin{equation}
\mathrm{\ell_{status}=\mathcal{F}_{status, 3}(u_{CLS}).}
\end{equation}
The logits $\mathrm{\ell_{status}}$ are normalized using softmax to yield a distribution over three possible statuses. During inference, the status with the highest probability is assigned to the slot.

\subsubsection{Categorical slot filler}

We apply a fully connected layer to each possible slot value embedding $\mathrm{u_{pv}}$ to obtain a logit:

\begin{equation}
\mathrm{\ell^j_{posval}=\mathcal{F}_{cat\_slot, 1}(\mathrm{u_{pv}}), 1 \leq j \leq m + 1},
\end{equation}
where $\mathrm{m}$ is the maximum number of possible categorical slot values. An additional value corresponds to the “NONE” value. The calculated logits are combined into a vector and normalized with softmax to get a distribution over all possible values.

\subsubsection{Free-form slot filler}

To get a span for a non-categorical slot value, we predict the span start and the span end distributions over token level representations $\mathrm{t_i}$ of dialogue history:
\begin{equation}
    \mathrm{\ell^i_{start}=\mathcal{F}_{start,1}(t_i), 1 \leq i \leq n},
\end{equation}
\begin{equation}
    \mathrm{\ell^j_{stop}=\mathcal{F}_{stop,1}(t_j), 1 \leq j \leq n},
\end{equation}
where $\mathrm{n}$ is equal to the hidden state dimension (typically 384 or 512, as required for BERT input).

\subsubsection{Requested slot gate}

A request for a slot value by the user is predicted by applying $F_{\mathrm{req\_slot}, 2}$ to the $\mathrm{u_{CLS}}$ embedding:
\begin{equation}
    \mathrm{\ell_{req\_slot}=\mathcal{F}_{req\_slot, 2}(u_{CLS}).}
\end{equation}

The calculated logits are normalized with softmax to yield a probability distribution over two possible requested slot statuses: \texttt{requested} or \texttt{not\_requested}. If the predicted status is \texttt{requested}, then the slot is added to the requested slots list.

\subsubsection{Intent classifier}

To predict the active user intent for a given service, we apply a fully connected layer to every $\mathrm{u_{int}}$ embedding and then obtain the probability distribution with softmax:
\begin{equation}
    \mathrm{\ell^j_{intent}=\mathcal{F}_{intent, 1}(u_{int}), 1 < j < s + 1,}
\end{equation}
where $\mathrm{s}$ is the maximum number of intents per service and the additional value corresponds to the “NONE” intent.


\section{Experimental Setup}

\subsection{Schema-Guided Dialogue Dataset}

To demonstrate the performance of our model, we use the recently released Schema-Guided Dialogue Dataset (SGD). It is the largest public task-oriented dialogue corpus as announced by its authors \cite{rastogi2019towards}. SGD incorporates 34 services related to 16 different domains with over 18,000 dialogues in both train and dev sets. The evaluation set contains unseen services and domains, so the model is expected to generalize in zero-shot settings. The dataset consists of single-domain and multi-domain dialogues. A single-domain dialogue involves interactions with only one service, while a multi-domain dialogue has interactions with two or more different services.

The authors also proposed the schema-guided approach for the task-oriented dialogue. A schema defines the interface for a backend API as a list of user intents and slots, as well as their natural language descriptions. Each dialogue in the dataset is accompanied by one or more schemas relevant to the dialogue (one schema corresponds to a single service). The model should use the service’s schema as input to produce predictions over the intents and slots listed in the schema. The natural language descriptions of slots and intents allow the model to handle unseen services.


\subsection{Training Details}

As an encoder, we use the pre-trained BERT model (bert-large-cased-whole-word-masking-finetuned-squad \footnote{https://huggingface.co/transformers/pretrained\_models.html}) with 24 layers of 1,024 hidden units, 16 self-attention heads and 340 million parameters. We fine-tune the model parameters using the Adam optimizer with weight decay \cite{loshchilov2018decoupled}. The initial learning rate of the optimizer was set to $3.5 \cdot 10^{-5}$. The total loss is defined as a sum of cross-entropy losses for each head. We train the model for 5 epochs with a batch size of 8 and 12 gradient accumulation steps on one Tesla V100 32GB.

Due to our training procedure we get a substantial amount of examples where a slot is not present in the state update and the model has to predict either an empty span or a “NONE” value. These instances (we term them “negative”) force the model to make constant predictions. In order to mitigate this issue, we introduce the \texttt{cat\_neg\_sampling\_prob} (by default 0.1) and \texttt{noncat\_neg\_sampling\_prob} (by default 0.2) sampling rates for categorical and non-categorical slots respectively. Also, the number of non-categorical examples overwhelms that of categorical ones. We deal with this class imbalance by providing separate batches for categorical and non-categorical examples.


\subsection{Evaluation}

The following metrics were used to evaluate the dialogue state tracking task:

\begin{itemize}
\item \textbf{Active Intent Accuracy}: The portion of user turns for which the active intent was correctly predicted.
\item \textbf{Requested Slot F1}: The macro-averaged F1 score for requested slots over all turns.
\item \textbf{Average Goal Accuracy}: For each user utterance, the model predicts a single value for every slot present in the dialogue state. Only the slots which have a non-empty assignment in the ground truth dialogue state are considered for this metric. This is the average accuracy of predicting the value of a slot correctly. A fuzzy matching score is used for non-categorical slots to reward partial matches with the ground truth.
\item \textbf{Joint Goal Accuracy}: This is the average accuracy of predicting all slot assignments for a turn correctly.
\end{itemize}


\section{Experimental Results}
\begin{figure}[t!]
	\resizebox{0.47 \textwidth}{!}{%
        \includegraphics[]{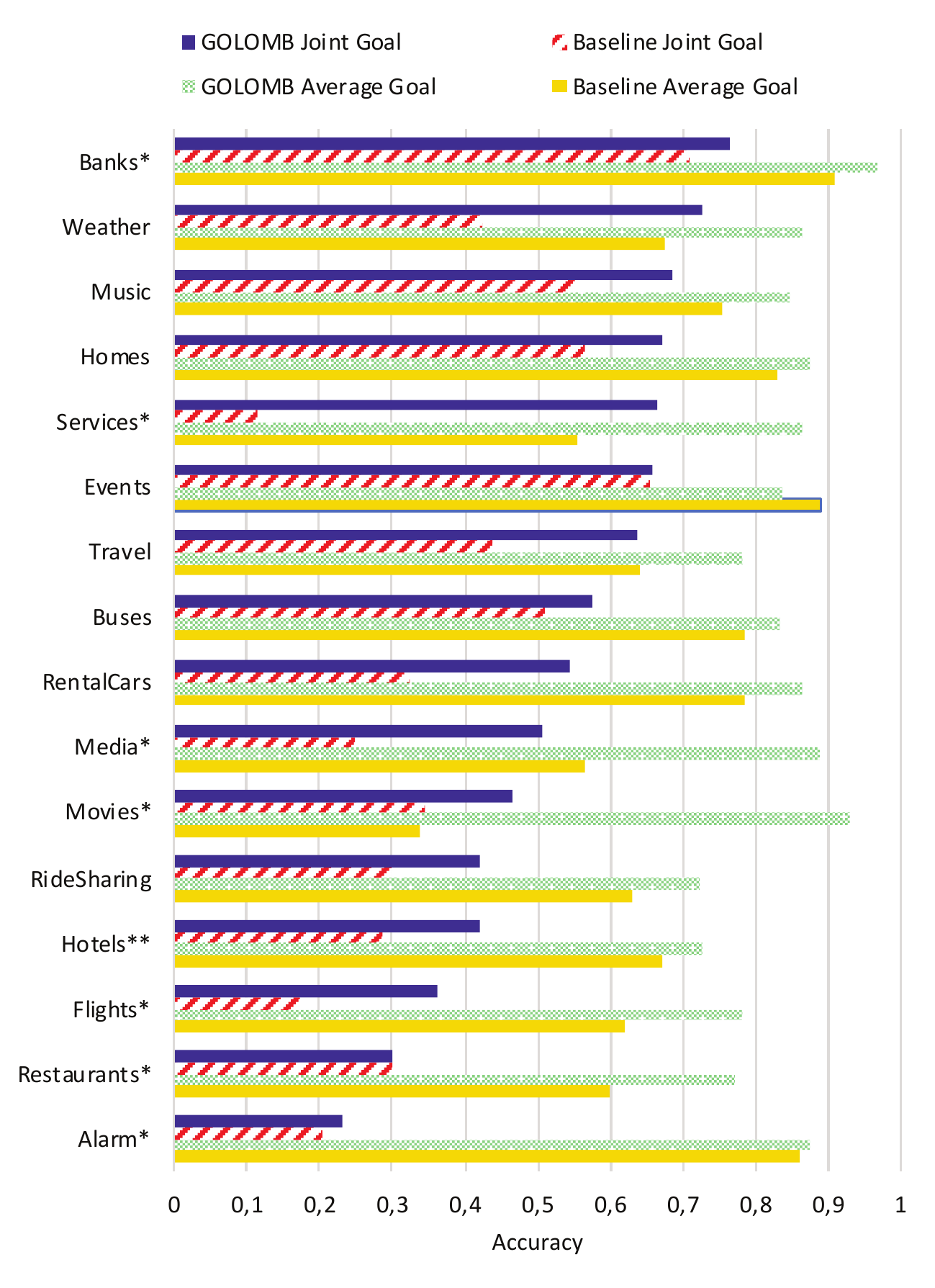}
	}
	\caption{
	Per-domain performance comparison by joint goal accuracy and average goal accuracy between the baseline and our model. Here, “*” denotes a domain with a service present in dev and not present in train, and “**” denotes the domain with one seen and one unseen service. The other domains contain services from train only.}
	\label{big_table}
\end{figure}

The results of GOLOMB evaluation on the dev and test sets and the dev scores of the baseline model are shown in Table \ref{results_table}. The comparison between our model and the baseline model across the different domains is provided in Figure \ref{big_table}.

As we can see from Table \ref{results_table}, our model outperforms the baseline model in terms of joint goal accuracy and average goal accuracy, whereas the baseline model has better scores for requested slot F1 and active intent accuracy. A plausible explanation for the significantly higher active intent score of the baseline model is that it uses the \texttt{[CLS]} token output for intent predictions. We also tried to employ the \texttt{[CLS]} token output for the intent classifier and got better scores for intent accuracy. But at the same time, joint goal accuracy degraded.

    \begin{table}[thb!]
	    \resizebox{0.47\textwidth}{!}{%
		    \begin{tabular}{c|c|c|c|c}
			    \Xhline{1pt}
                & \textbf{Active Int Acc} & \textbf{Req Slot F1} & \textbf{Avg GA} & \textbf{Joint GA} \\
                \hline
                \textbf{GOLOMB, dev scores} & 0.660 & 0.969 & \textbf{0.817} & \textbf{0.539} \\
                \hline
                \textbf{Baseline, dev scores} & \textbf{0.908} & \textbf{0.973} & 0.740 & 0.411 \\
                \hline
                \hline
                \textbf{GOLOMB, test scores} & 0.747 & 0.971 & 0.750 & 0.465 \\
                \Xhline{1pt}
		    \end{tabular}
        }
   \caption{Performance comparison between the baseline and our model on the dev set, and our model’s scores on the test set.}
   \label{results_table}
\end{table}

\begin{figure}[tbh!]
	\resizebox{0.47 \textwidth}{!}{%
        \includegraphics[]{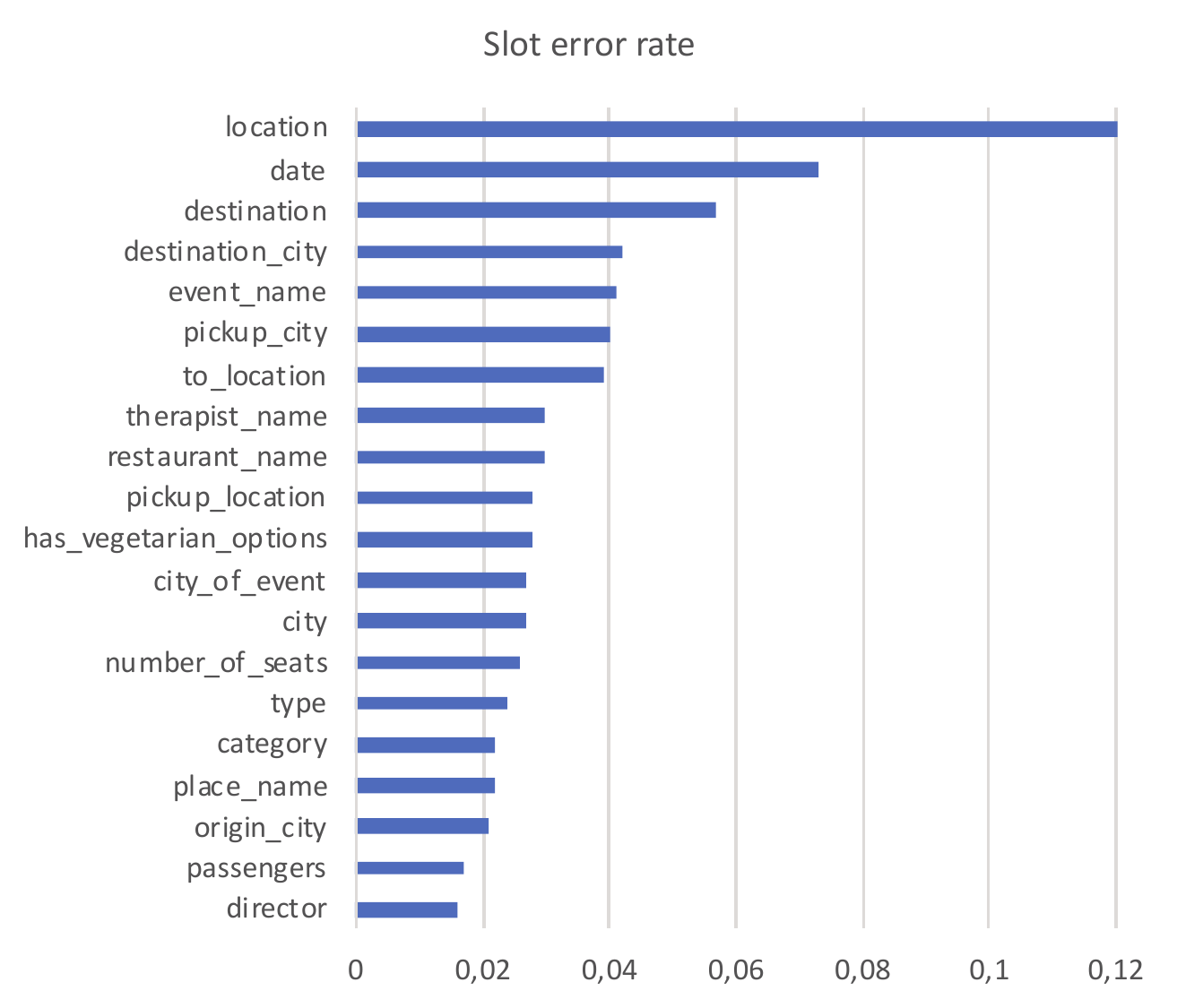}
	}
	\caption{
		First 20 slots sorted by the error rate on the dev set. The \emph{location} slot, which appears in the \emph{Hotels}, \emph{Restaurants} and \emph{Travel} domains, has the highest error rate, 12\%. The \emph{director} slot from the \emph{Media} and \emph{Movies} domains has the lowest error rate, 1.6\%.}
	\label{error_rate}
\end{figure}

Figure \ref{big_table} shows a comparison between our model and the baseline model by joint goal accuracy and average goal accuracy. Our model exhibits better performance in every domain by joint goal accuracy. But in the \emph{Events} domain both models performed well due to the large number of training examples in that domain. The greatest gap is evident in the \emph{Services} domain, where our model shows superior performance. As one can observe from Figure \ref{big_table}, the model’s performance is in general better on the domains for which services were present in the train set. However, the best result for joint goal accuracy is achieved on the \emph{Banks} domain, even though the corresponding service was unseen during training.

The \emph{Alarm} domain exhibits the worst performance by joint goal accuracy. The most likely explanation is that no examples from this domain were seen by the model during training. However, a relatively good performance by average accuracy signals the presence of a few especially deceptive slots, on which the model makes more mistakes than elsewhere.

\begin{figure}[t!]
	\resizebox{0.475\textwidth}{!}{%
        \includegraphics[]{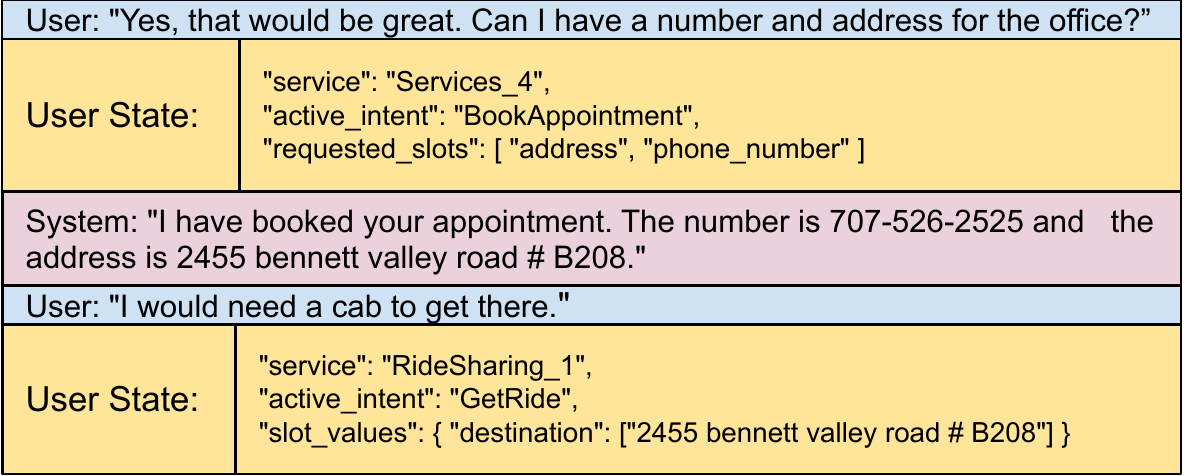}
	}
	\caption{An example of domain switch in the dialogue 13\_00089 from the dev set. The user requested a slot \emph{address} from the \emph{Services} domain, but its value was assigned to the \emph{destination} slot from the \emph{RideSharing} domain.}
	\label{example}
\end{figure}
\begin{table*}[t!]
	    \resizebox{\textwidth}{!}{%
		    \begin{tabular}{c|c|c|c|c|c||c|c|c|c}
			    \Xhline{1pt}
                & NLD & CLS for categorical slots & PV for categorical slots & Intents &  SQuAD pre-trainig & \textbf{Active Int Acc} & \textbf{Req Slot F1} & \textbf{Avg GA} & \textbf{Joint GA} \\
                \hline
                \textbf{(a)} & -- & + & -- & -- & -- & -- & 0.969 & 0.782 & 0.460 \\
                \hline
                \textbf{(b)} & + & + & -- & -- & -- & -- & 0.969 & 0.778 & 0.464 \\
                \hline
                \textbf{(c)} & + & -- & + & -- & -- & -- & 0.969 & 0.814 & 0.524 \\
                \hline
                \textbf{(d)} & + & -- & + & + & -- & 0.657 & 0.969 & \textbf{0.820} & 0.535 \\
                \hline
                \textbf{Final model} & + & -- & + & + & + & \textbf{0.660} & 0.969 & 0.817 & \textbf{0.539} \\
                \Xhline{1pt}
		    \end{tabular}
        }
   \caption{Ablation study. Here, “NLD” denotes the case when the natural language descriptions of slots and domains are used, “PV” denotes possible slot values. For categorical slot value prediction, two approaches were implemented. The first approach, described in detail below, is to use the $\mathrm{u_{CLS}}$ output. The second approach, which is part of our final architecture, uses the outputs $\mathrm{u_{pv}}$ of special $\mathrm{pv}$ tokens to select a slot value among the possible values.}
   \label{ablation_table}
\end{table*}

The error rate across the different slots is shown in Figure \ref{error_rate}. Not surprisingly, the \emph{location} slot has the highest error rate (12\%), as it appears in three domains: \emph{Hotels}, \emph{Restaurants} and \emph{Travel}, of which only the \emph{Travel} domain was seen during training. The \emph{date} slot has the second-highest error rate of 7\% and also appears in three domains, of which the \emph{Restaurants} domain was unseen. The slots \emph{destination} and \emph{destination\_city} also have high error rates of 6\% and 4\% respectively. These slots were often filled with origin places instead of destination places. However, the mismatch of origin and destination points was a frequent mistake in the SGD dataset itself, so the model could be confused by incorrect labels.

In multi-domain dialogues, we noticed that our model frequently makes mistakes on the turns where a domain switch happens. Typically, a false state tracking happens in a situation when a slot has been tracked for one domain and its value needs to be transferred to a slot in the new domain. The main obstacle here is that there is no mention of the value for the new slot in the recent context and our model cannot find this slot’s value by design.

An example of such a situation can be found in the dialogue 13\_00089 from the dev set (see Figure \ref{example}). Here, the user requested a slot \emph{address} which corresponds to the domain \emph{Services}, but its value was assigned to the \emph{destination} slot from the \emph{RideSharing} domain. Our model is not able to share slot values directly between different domains, so the slot \emph{destination} was filled incorrectly with the word “there” (the ending of the last user utterance).

\subsection{Ablation Study}

The results of an ablation study for our model are provided in Table \ref{ablation_table}. We perform the following ablation experiments:
\begin{enumerate}[(a)]
    \item \textbf{CLS for categorical slots.} Our first version of the categorical slot filler used the $\mathrm{u_{CLS}}$ output to predict categorical slot values. $\mathrm{u_{CLS}}$ was fed into a fully connected layer with $\mathrm{m} + 1$ outputs, where $\mathrm{m}$ is the maximum number of possible categorical slot values. The last position always corresponded to the “NONE” value. If a slot had $\mathrm{k} < \mathrm{m}$ possible values, the positions between $\mathrm{k + 1}$ and $\mathrm{m + 1}$ were filled with --INF to get zero probabilities after applying softmax. We also fed only the slot and domain names to the BERT Encoder, without their natural language descriptions.
    \item \textbf{CLS for categorical slots + NLD.} We added natural language descriptions (NLD) to the previous setup. Surprisingly, the increment in performance was not as substantial as we expected.
    \item \textbf{PV for categorical slots + NLD.} Introducing special $\mathrm{pv}$ tokens for possible values led to a huge increase of around 6\% in performance by joint goal accuracy.
    \item \textbf{PV for categorical slots + NLD + Intents.} We implemented intent prediction by introducing special $\mathrm{u_{int}}$ tokens for possible user intents (in the same manner as for categorical slots). Though the intent prediction accuracy is not particularly high, the overall performance showed an increase of around 1\% by joint goal accuracy.
    \item \textbf{PV for categorical slots + NLD + Intents + SQuAD pre-training.} The encoder we used for our final model was the BERT model fine-tuned on the SQuAD dataset. We got an increase in joint goal accuracy, so we did not give up SQuAD pre-training, even though the average goal accuracy deteriorated.
\end{enumerate}
\section{Conclusion}
We proposed a multi-task BERT-based model for multi-domain dialogue state tracking in zero-shot settings. Our approach is robust to schema modifications and is able to transfer the extracted knowledge to unseen domains. The model is consistent with real-life scenarios raised by virtual assistants and achieves substantial improvements over the baseline.


\section{Acknowledgments}

The work was supported by National Technology Initiative and PAO Sberbank project ID
0000000007417F630002.

\bibliography{gulyaev.bib}

\begin{thebibliography}{}

\bibitem[\protect\citeauthoryear{Budzianowski \bgroup et al\mbox.\egroup
  }{2018}]{budzianowski2018multiwoz}
Budzianowski, P.; Wen, T.-H.; Tseng, B.-H.; Casanueva, I.; Ultes, S.; Ramadan,
  O.; and Ga{\v{s}}i{\'c}, M.
\newblock 2018.
\newblock {M}ulti{WOZ} - a large-scale multi-domain wizard-of-{O}z dataset for
  task-oriented dialogue modelling.
\newblock In {\em Proceedings of the 2018 Conference on Empirical Methods in
  Natural Language Processing},  5016--5026.
\newblock Brussels, Belgium: Association for Computational Linguistics.

\bibitem[\protect\citeauthoryear{Chao and Lane}{2019}]{chao2019bert}
Chao, G.-L., and Lane, I.
\newblock 2019.
\newblock {BERT-DST}: Scalable end-to-end dialogue state tracking with
  bidirectional encoder representations from transformer.
\newblock In {\em INTERSPEECH}.

\bibitem[\protect\citeauthoryear{Devlin \bgroup et al\mbox.\egroup
  }{2019}]{devlin2018bert}
Devlin, J.; Chang, M.-W.; Lee, K.; and Toutanova, K.
\newblock 2019.
\newblock Bert: Pre-training of deep bidirectional transformers for language
  understanding.
\newblock In {\em North American Association for Computational Linguistics
  (NAACL)}.

\bibitem[\protect\citeauthoryear{Eric \bgroup et al\mbox.\egroup
  }{2019}]{eric2019multiwoz}
Eric, M.; Goel, R.; Paul, S.; Sethi, A.; Agarwal, S.; Gao, S.; and Hakkani-Tur,
  D.
\newblock 2019.
\newblock Multiwoz 2.1: Multi-domain dialogue state corrections and state
  tracking baselines.
\newblock {\em arXiv preprint arXiv:1907.01669}.

\bibitem[\protect\citeauthoryear{Goel, Paul, and
  Hakkani-Tür}{2019}]{goel2019hyst}
Goel, R.; Paul, S.; and Hakkani-Tür, D.
\newblock 2019.
\newblock {HyST: A Hybrid Approach for Flexible and Accurate Dialogue State
  Tracking}.
\newblock In {\em Proc. Interspeech 2019},  1458--1462.

\bibitem[\protect\citeauthoryear{Hemphill, Godfrey, and
  Doddington}{1990}]{hemphill1990atis}
Hemphill, C.~T.; Godfrey, J.~J.; and Doddington, G.~R.
\newblock 1990.
\newblock The atis spoken language systems pilot corpus.
\newblock {\em Speech and Natural Language: Proceedings of a Workshop Held at
  Hidden Valley, Pennsylvania, June 24-27, 1990}.

\bibitem[\protect\citeauthoryear{Henderson, Thomson, and
  Williams}{2014}]{henderson-etal-2014-second}
Henderson, M.; Thomson, B.; and Williams, J.~D.
\newblock 2014.
\newblock The second dialog state tracking challenge.
\newblock In {\em Proceedings of the 15th Annual Meeting of the Special
  Interest Group on Discourse and Dialogue ({SIGDIAL})},  263--272.
\newblock Philadelphia, PA, U.S.A.: Association for Computational Linguistics.

\bibitem[\protect\citeauthoryear{Henderson, Thomson, and
  Young}{2014}]{henderson2014word}
Henderson, M.; Thomson, B.; and Young, S.
\newblock 2014.
\newblock Word-based dialog state tracking with recurrent neural networks.
\newblock In {\em Proceedings of the 15th Annual Meeting of the Special
  Interest Group on Discourse and Dialogue (SIGDIAL)},  292--299.

\bibitem[\protect\citeauthoryear{Konovalov \bgroup et al\mbox.\egroup
  }{2016}]{konovalov2016negochat}
Konovalov, V.; Artstein, R.; Melamud, O.; and Dagan, I.
\newblock 2016.
\newblock The negochat corpus of human-agent negotiation dialogues.
\newblock In {\em Proceedings of the Tenth International Conference on Language
  Resources and Evaluation (LREC'16)}.

\bibitem[\protect\citeauthoryear{Loshchilov and
  Hutter}{2018}]{loshchilov2018decoupled}
Loshchilov, I., and Hutter, F.
\newblock 2018.
\newblock Decoupled weight decay regularization.

\bibitem[\protect\citeauthoryear{Mrk{\v{s}}i{\'c} \bgroup et al\mbox.\egroup
  }{2017}]{mrkvsic2017neural}
Mrk{\v{s}}i{\'c}, N.; S{\'e}aghdha, D.~O.; Wen, T.-H.; Thomson, B.; and Young,
  S.
\newblock 2017.
\newblock Neural belief tracker: Data-driven dialogue state tracking.
\newblock In {\em Proceedings of the 55th Annual Meet- ing of the Association
  for Computational Linguistics (Volume 1: Long Papers)},  1777--1788.
\newblock Association for Computational Linguistics.

\bibitem[\protect\citeauthoryear{Nouri and
  Hosseini-Asl}{2018}]{nouri2018toward}
Nouri, E., and Hosseini-Asl, E.
\newblock 2018.
\newblock Toward scalable neural dialogue state tracking model.
\newblock {\em Advances in neural information processing systems (NeurIPS), 2nd
  Conversational AI workshop}.

\bibitem[\protect\citeauthoryear{Rajpurkar \bgroup et al\mbox.\egroup
  }{2016}]{rajpurkar2016squad}
Rajpurkar, P.; Zhang, J.; Lopyrev, K.; and Liang, P.
\newblock 2016.
\newblock Squad: 100,000+ questions for machine comprehension of text.
\newblock In {\em Proceedings of the 2016 Conference on Empirical Methods in
  Natural Language Processing},  2383--2392.

\bibitem[\protect\citeauthoryear{Rastogi \bgroup et al\mbox.\egroup
  }{2019}]{rastogi2019towards}
Rastogi, A.; Zang, X.; Sunkara, S.; Gupta, R.; and Khaitan, P.
\newblock 2019.
\newblock Towards scalable multi-domain conversational agents: The
  schema-guided dialogue dataset.
\newblock {\em arXiv preprint arXiv:1909.05855}.

\bibitem[\protect\citeauthoryear{Ren \bgroup et al\mbox.\egroup
  }{2018}]{ren2018towards}
Ren, L.; Xie, K.; Chen, L.; and Yu, K.
\newblock 2018.
\newblock Towards universal dialogue state tracking.
\newblock In {\em Proceedings of the 2018 Conference on Empirical Methods in
  Natural Language Processing},  2780--2786.

\bibitem[\protect\citeauthoryear{Thomson and Young}{2010}]{thomson2010bayesian}
Thomson, B., and Young, S.
\newblock 2010.
\newblock Bayesian update of dialogue state: A pomdp framework for spoken
  dialogue systems.
\newblock {\em Computer Speech \& Language} 24(4):562--588.

\bibitem[\protect\citeauthoryear{Wang and Lemon}{2013}]{wang2013simple}
Wang, Z., and Lemon, O.
\newblock 2013.
\newblock A simple and generic belief tracking mechanism for the dialog state
  tracking challenge: On the believability of observed information.
\newblock In {\em Proceedings of the SIGDIAL 2013 Conference},  423--432.

\bibitem[\protect\citeauthoryear{Wen \bgroup et al\mbox.\egroup
  }{2017}]{wen2016network}
Wen, T.-H.; Vandyke, D.; Mrksic, N.; Gasic, M.; Rojas-Barahona, L.~M.; Su,
  P.-H.; Ultes, S.; and Young, S.
\newblock 2017.
\newblock A network-based end-to-end trainable task-oriented dialogue system.
\newblock In {\em Proceedings of the 15th Conference of Bayesian the European
  Chapter of the Association for Computational Linguistics: Volume 1, Long
  Papers},  292--299.
\newblock Association for Computational Linguistics.

\bibitem[\protect\citeauthoryear{Williams}{2014}]{williams2014web}
Williams, J.~D.
\newblock 2014.
\newblock Web-style ranking and slu combination for dialog state tracking.
\newblock In {\em Proceedings of the 15th Annual Meeting of the Special
  Interest Group on Discourse and Dialogue (SIGDIAL)},  282--291.

\bibitem[\protect\citeauthoryear{Wu \bgroup et al\mbox.\egroup
  }{2019}]{wu2019transferable}
Wu, C.-S.; Madotto, A.; Hosseini-Asl, E.; Xiong, C.; Socher, R.; and Fung, P.
\newblock 2019.
\newblock Transferable multi-domain state generator for task-oriented dialogue
  systems.
\newblock In {\em Proceedings of the 57th Annual Meeting of the Association for
  Computational Linguistics (Volume 1: Long Papers)}.
\newblock Association for Computational Linguistics.

\bibitem[\protect\citeauthoryear{Xu and Hu}{2018}]{xu2018end}
Xu, P., and Hu, Q.
\newblock 2018.
\newblock An end-to-end approach for handling unknown slot values in dialogue
  state tracking.
\newblock In {\em Proceedings of the 56th Annual Meet- ing of the Association
  for Computational Linguistics (Volume 1: Long Papers)},  1448--1457.
\newblock Association for Computational Linguistics.

\bibitem[\protect\citeauthoryear{Zhong, Xiong, and
  Socher}{2018}]{zhong2018global}
Zhong, V.; Xiong, C.; and Socher, R.
\newblock 2018.
\newblock Global-locally self-attentive encoder for dialogue state tracking.
\newblock In {\em Proceedings of the 56th Annual Meeting of the Association for
  Computational Linguistics (Volume 1: Long Papers)},  1458--1467.

\bibitem[\protect\citeauthoryear{Zilka and
  Jurcicek}{2015}]{zilka2015incremental}
Zilka, L., and Jurcicek, F.
\newblock 2015.
\newblock Incremental lstm-based dialog state tracker.
\newblock In {\em 2015 Ieee Workshop on Automatic Speech Recognition and
  Understanding (Asru)},  757--762.
\newblock IEEE.

\end{thebibliography}
\bibliographystyle{aaai}
\end{document}